\documentclass[11pt,a4paper]{article}
\usepackage[hyperref]{aacl-ijcnlp2020}
\usepackage{times}
\usepackage{latexsym}
\usepackage{mathrsfs}
\usepackage{amsmath}
\usepackage{amssymb}
\usepackage{booktabs}
\usepackage{graphicx}
\usepackage{multirow}
\usepackage{makecell}
\usepackage{hyperref}
\usepackage[toc,page]{appendix} 

\usepackage{microtype}

\aclfinalcopy 
\def\aclpaperid{276} 

\title{IndoNLU: Benchmark and Resources for Evaluating Indonesian \\Natural Language Understanding}

\author{Bryan Wilie$^{1}$\thanks{\hspace{1mm} These authors contributed equally.} , Karissa Vincentio$^{2*}$, Genta Indra Winata$^{3*}$, Samuel Cahyawijaya$^{3*}$, \\
\textbf{Xiaohong Li}$^4$, \textbf{Zhi Yuan Lim}$^4$, \textbf{Sidik Soleman}$^{5}$, \textbf{Rahmad Mahendra}$^6$, \\
\textbf{Pascale Fung}$^{3}$, \textbf{Syafri Bahar}$^{4}$, \textbf{Ayu Purwarianti}$^{1,5}$ \\
$^1$Institut Teknologi Bandung \hspace{5mm}
$^2$Universitas Multimedia Nusantara \\
$^3$The Hong Kong University of Science and Technology\\
$^4$Gojek\hspace{5mm} $^5$Prosa.ai \hspace{5mm}
$^6$Universitas Indonesia\\
\small\texttt{\{bryanwilie92, karissavin\}@gmail.com, \{giwinata, scahyawijaya\}@connect.ust.hk}}

\date{}

\begin{document}
\maketitle
\begin{abstract}
Although Indonesian is known to be the fourth most frequently used language over the internet, the research progress on this language in natural language processing (NLP) is slow-moving due to a lack of available resources. In response, we introduce the first-ever vast resource for training, evaluation, and benchmarking on Indonesian natural language understanding (\texttt{IndoNLU}) tasks. \texttt{IndoNLU} includes twelve tasks, ranging from single sentence classification to pair-sentences sequence labeling with different levels of complexity. The datasets for the tasks lie in different domains and styles to ensure task diversity. We also provide a set of Indonesian pre-trained models (IndoBERT) trained from a large and clean Indonesian dataset (\texttt{Indo4B}) collected from publicly available sources such as social media texts, blogs, news, and websites. We release baseline models for all twelve tasks, as well as the framework for benchmark evaluation, thus enabling everyone to benchmark their system performances.
\end{abstract}

\section{Introduction}
\label{sec:intro}

Following the notable success of contextual pre-trained language methods~\cite{peters2018deep,devlin2019bert}, several benchmarks to gauge the progress of general-purpose NLP research, such as GLUE~\cite{wang2018glue}, SuperGLUE~\cite{wang2019superglue}, and CLUE~\cite{xu2020clue}, have been proposed. These benchmarks cover a large range of tasks to measure how well pre-trained models achieve compared to humans. However, these metrics are limited to high-resource languages, such as English and Chinese, that already have existing datasets available and are accessible to the research community. Most languages, by contrast, suffer from limited data collection and low awareness of published data for research. One of the languages which suffer from this resource scarcity problem is Indonesian.

Indonesian is the fourth largest language used over the internet, with around 171 million users across the globe.\footnote{https://www.internetworldstats.com/stats3.htm} 
Despite a large amount of Indonesian data available over the internet, the advancement of NLP research in Indonesian is slow-moving. This problem occurs because available datasets are scattered, with a lack of documentation and minimal community engagement. Moreover, many existing studies in Indonesian NLP do not provide codes and test splits, making it impossible to reproduce results.

To address the data scarcity problem, we propose the first-ever Indonesian natural language understanding benchmark, \texttt{IndoNLU}, a collection of twelve diverse tasks. The tasks are mainly categorized based on the input, such as single-sentences and sentence-pairs, and objectives, such as sentence classification tasks and sequence labeling tasks. The benchmark is designed to cater to a range of styles in both formal and colloquial Indonesian, which are highly diverse. We collect a range of datasets from existing works: an emotion classification dataset ~\cite{saputri2018emotion}, QA factoid dataset ~\cite{purwarianti2007qa}, sentiment analysis dataset ~\cite{purwarianti2019improving}, aspect-based sentiment analysis dataset ~\cite{ilmania2018aspect, azhar2019aspectcategorization}, part-of-speech (POS) tag dataset ~\cite{Dinakaramani2014, hoesen2018investigating}, named entity recognition (NER) dataset ~\cite{hoesen2018investigating}, span extraction dataset ~\cite{mahfuzh2019improving, septi2019aspect, fern2019aspect},  and textual entailment dataset ~\cite{ken2018entailment}. It is difficult to compare model performance since there is no official split of information for existing datasets. Therefore we standardize the benchmark by resplitting the datasets on each task for reproducibility purposes. To expedite the modeling and evaluation processes for this benchmark, we present samples of the model pre-training code and a framework to evaluate models in all downstream tasks. We will publish the score of our benchmark on a publicly accessible leaderboard to provide better community engagement and benchmark transparency.

To further advance Indonesian NLP research, we collect around four billion words from Indonesian preprocessed text data ($\approx$ 23 GB), as a new standard dataset, called \texttt{Indo4B}, for self-supervised learning. The dataset comes from sources like online news, social media, Wikipedia, online articles, subtitles from video recordings, and parallel datasets. We then introduce an Indonesian BERT-based model, IndoBERT, which is trained on our \texttt{Indo4B} dataset. We also introduce another IndoBERT variant based on the ALBERT model \cite{Lan2020ALBERT}, called IndoBERT-lite. The two variants of IndoBERT are used as baseline models in the \texttt{IndoNLU} benchmark. In this work, we also extensively compare our IndoBERT models to different pre-trained word embeddings and existing multilingual pre-trained models, such as Multilingual BERT~\cite{devlin2019bert} and XLM-R~\cite{conneau2019xlmr}, to measure their effectiveness. Results show that our pre-trained models outperform most of the existing pre-trained models.

\section{Related Work}
\paragraph{Benchmarks} GLUE~\cite{wang2018glue} is a multi-task benchmark for natural language understanding (NLU) in the English language. It consists of nine tasks: single-sentence input, semantic similarity detection, and natural language inference (NLI) tasks. GLUE's harder counterpart SuperGLUE ~\cite{wang2019superglue} covers question answering, NLI, co-reference resolution, and word sense disambiguation tasks. CLUE~\cite{xu2020clue} is a Chinese NLU benchmark that includes a test set designed to probe a unique and specific linguistic phenomenon in the Chinese language. It consists of eight diverse tasks, including single-sentence, sentence-pair, and machine reading comprehension tasks. FLUE~\cite{le2019flaubert} is an evaluation NLP benchmark for the French language which is divided into six different task categories: text classification, paraphrasing, NLI, parsing, POS tagging, and word sense disambiguation.

\paragraph{Contextual Language Models} 
In recent years, contextual pre-trained language models have shown a major breakthrough in NLP, starting from ELMo~\cite{peters2018deep}. With the emergence of the transformer model~\cite{vaswani2017attention}, ~\citet{devlin2019bert} proposed BERT, a faster architecture to train a language model that eliminates recurrences by applying a multi-head attention layer. \citet{liu2019roberta} later proposed RoBERTa, which improves the performance of BERT by applying dynamic masking, increasing the batch size, and removing the next-sentence prediction. ~\citet{Lan2020ALBERT} proposed ALBERT, which extends the BERT model by applying factorization and weight sharing to reduce the number of parameters and time.

Many research studies have introduced contextual pre-trained language models on languages other than English. \citet{cui2019pretraining} introduced the Chinese BERT and RoBERTa models, while \citet{martin2019camembert} and \citet{le2019flaubert} introduced CamemBERT and FLAUBert respectively, which are BERT-based models for the French language. \citet{devlin2019bert} introduced the Multilingual BERT model, a BERT model trained on monolingual Wikipedia data in many languages. Meanwhile, \citet{lample2019xlm} introduced XLM, a cross-lingual pre-trained language model that uses parallel data as a new translation masked loss to improve the cross-linguality. Finally, \citet{conneau2019xlmr} introduced XLM-R, a RoBERTa-based XLM model.

\begin{table*}[!t]
\centering
\resizebox{0.99\textwidth}{!}{
\begin{tabular}{lccclccll}
\toprule
\textbf{Dataset} & $|$\textbf{Train}$|$ & $|$\textbf{Valid}$|$ & $|$\textbf{Test}$|$ & \textbf{Task Description} & \textbf{\#Label}  & \textbf{\#Class} & \textbf{Domain} & \textbf{Style}\\ \midrule
\multicolumn{8}{c}{Single-Sentence Classification Tasks} \\ \midrule
EmoT$^{\dagger}$ & 3,521 & 440 & 442 & emotion classification & 1 & 5 & tweets & colloquial \\ 
SmSA & 11,000 & 1,260 & 500 & sentiment analysis & 1 & 3 & general & colloquial \\ 
CASA & 810 & 90 & 180 & aspect-based sentiment analysis & 6 & 3 & automobile & colloquial \\
HoASA$^{\dagger}$ & 2,283 & 285 & 286 & aspect-based sentiment analysis & 10 & 4 & hotel &  colloquial \\ \midrule
\multicolumn{8}{c}{Sentence-Pair Classification Tasks} \\ \midrule
WReTE$^{\dagger}$ & 300 & 50 & 100 & textual entailment & 1 & 2 & wiki & formal \\ \midrule
\multicolumn{8}{c}{Single-Sentence Sequence Labeling Tasks} \\ \midrule
POSP$^{\dagger}$ & 6,720 & 840 & 840 & part-of-speech tagging & 1 & 26 & news & formal \\ 
BaPOS & 8,000 & 1,000 & 1,029 & part-of-speech tagging & 1 & 41 & news & formal \\
TermA & 3,000 & 1,000 & 1,000 & span extraction & 1 & 5 & hotel & colloquial \\
KEPS & 800 & 200 & 247 & span extraction & 1 & 3 & banking & colloquial \\
NERGrit$^{\dagger}$ & 1,672 & 209 & 209 & named entity recognition & 1 & 7 & wiki & formal \\
NERP$^{\dagger}$ & 6,720 & 840 & 840 & named entity recognition & 1 & 11 & news & formal \\ \midrule
\multicolumn{8}{c}{Sentence-Pair Sequence Labeling Tasks} \\ \midrule FacQA & 2,495 & 311 & 311 & span extraction  & 1 & 3 & news & formal \\ \bottomrule
\end{tabular}
}
\caption{Task statistics and descriptions. $^{\dagger}$We create new splits for the dataset.}
\label{tab:dataset}
\end{table*}

\section{\texttt{IndoNLU} Benchmark}
\label{sec:indonlu}

In this section, we describe our benchmark as four components. Firstly, we introduce the 12 tasks in \texttt{IndoNLU} for Indonesian natural language understanding. Secondly, we introduce a large-scale Indonesian dataset for self-supervised pre-training models. Thirdly, we explain the various kinds of baseline models used in our \texttt{IndoNLU} benchmark. Lastly, we describe the evaluation metric used to standardize the scoring over different models in our \texttt{IndoNLU} benchmark.


\subsection{Downstream Tasks}

The \texttt{IndoNLU} downstream tasks covers 12 tasks divided into four categories: (a) single-sentence classification, (b) single-sentence sequence-tagging, (c) sentence-pair classification, and (d) sentence-pair sequence labeling. The data samples for each task are shown in Appendix A.

\subsubsection{Single-Sentence Classification Tasks}
\paragraph{EmoT} An emotion classification dataset collected from the social media platform Twitter ~\cite{saputri2018emotion}. The dataset consists of around 4000 Indonesian colloquial language tweets, covering five different emotion labels: anger, fear, happiness, love, and sadness. 

\paragraph{SmSA} This sentence-level sentiment analysis dataset ~\cite{purwarianti2019improving} is a collection of comments and reviews in Indonesian obtained from multiple online platforms.
The text was crawled and then annotated by several Indonesian linguists to construct this dataset. There are three possible sentiments on the SmSA dataset: positive, negative, and neutral.

\paragraph{CASA} An aspect-based sentiment analysis dataset consisting of around a thousand car reviews collected from multiple Indonesian online automobile platforms~\cite{ilmania2018aspect}. The dataset covers six aspects of car quality. We define the task to be a multi-label classification task, where each label represents a sentiment for a single aspect with three possible values: positive, negative, and neutral.

\paragraph{HoASA} An aspect-based sentiment analysis dataset consisting of hotel reviews collected from the hotel aggregator platform, AiryRooms ~\cite{azhar2019aspectcategorization}.\footnote{\hyperlink{https://github.com/annisanurulazhar/absa-playground}{https://github.com/annisanurulazhar/absa-playground}} The dataset covers ten different aspects of hotel quality. Similar to the CASA dataset, each review is labeled with a single sentiment label for each aspect. There are four possible sentiment classes for each sentiment label: positive, negative, neutral, and positive-negative. The positive-negative label is given to a review that contains multiple sentiments of the same aspect but for different objects (e.g., cleanliness of bed and toilet). 

\begin{table*}[!t]
\centering
\resizebox{0.92\textwidth}{!}{
\begin{tabular}{lrccrrrcc}
\toprule
\textbf{Model} & \textbf{\#Params} & \textbf{\#Layers} &  \textbf{\#Heads} & \makecell{\textbf{Emb.} \\ \textbf{Size}} &  \makecell{\textbf{Hidden} \\ \textbf{Size}} & \makecell{\textbf{FFN} \\ \textbf{Size}} & \makecell{\textbf{Language} \\ \textbf{Type}} & \makecell{\textbf{Pre-train} \\ \textbf{Emb. Type}} \\ 
\midrule
Scratch & 15.1M & 6 & 10 & 300 & 300 & 3072 & Mono & - \\ \midrule
fastText-cc-id & 15.1M & 6 & 10 & 300 & 300 & 3072 & Mono & Word Emb. \\
fastText-indo4b & 15.1M & 6 & 10 & 300 & 300 & 3072 & Mono & Word Emb. \\
\midrule
$\text{IndoBERT-lite}_{\text{BASE}}$ & 11.7M & 12 & 12 & 128 & 768 & 3072 & Mono & Contextual \\
$\text{IndoBERT}_{\text{BASE}}$ & 124.5M & 12 & 12 & 768 & 768 & 3072 & Mono & Contextual \\
$\text{IndoBERT-lite}_{\text{LARGE}}$ &  17.7M & 24 & 16 & 128 & 1024 & 4096 & Mono & Contextual \\
$\text{IndoBERT}_{\text{LARGE}}$ & 335.2M & 24 & 16 & 1024 & 1024 & 4096 & Mono & Contextual \\
\midrule
mBERT & 167.4M & 12 & 12 & 768 & 768 & 3072 & Multi & Contextual \\
$\text{XLM-R}_{\text{BASE}}$ & 278.7M & 12 & 12 & 768 & 768 & 3072 & Multi & Contextual \\
$\text{XLM-R}_{\text{LARGE}}$ & 561.0M & 24 & 16  & 1024 & 1024 & 4096 & Multi & Contextual \\
$\text{XLM-MLM}_{\text{LARGE}}$ & 573.2M & 16 & 16 & 1280 & 1280 & 5120 & Multi & Contextual \\
\bottomrule
\end{tabular}
}
\caption{The details of baseline models used in \texttt{IndoNLU} benchmark}
\label{tab:baseline-models}
\end{table*}

\subsubsection{Sentence-Pair Classification Task}
\paragraph{WReTE}
The Wiki Revision Edits Textual Entailment dataset ~\cite{ken2018entailment} consists of 450 sentence pairs constructed from Wikipedia revision history. The dataset contains pairs of sentences and binary semantic relations between the pairs. The data are labeled as entailed when the meaning of the second sentence can be derived from the first one, and not entailed otherwise.

\subsubsection{Single-Sentence Sequence Labeling Tasks}
\paragraph{POSP} This Indonesian part-of-speech tagging (POS) dataset ~\cite{hoesen2018investigating} is collected from Indonesian news websites. The dataset consists of around 8000 sentences with 26 POS tags. The POS tag labels follow the Indonesian Association of Computational Linguistics (INACL) POS Tagging Convention.\footnote{\hyperlink{http://inacl.id/inacl/wp-content/uploads/2017/06/INACL-POS-Tagging-Convention-26-Mei.pdf}{http://inacl.id/inacl/wp-content/uploads/2017/06/INACL-POS-Tagging-Convention-26-Mei.pdf}}

\paragraph{BaPOS} This POS tagging dataset~\cite{Dinakaramani2014} contains about 1000 sentences, collected from the PAN Localization Project.\footnote{\hyperlink{http://www.panl10n.net/}{http://www.panl10n.net/}} In this dataset, each word is tagged by one of 23 POS tag classes.\footnote{\hyperlink{http://bahasa.cs.ui.ac.id/postag/downloads/Tagset.pdf}{http://bahasa.cs.ui.ac.id/postag/downloads/Tagset.pdf}} Data splitting used in this benchmark follows the experimental setting used by~\citet{kurniawan2018toward}.

\paragraph{TermA} This span-extraction dataset is collected from the hotel aggregator platform, AiryRooms ~\cite{septi2019aspect, fern2019aspect}.\footnote{\hyperlink{https://github.com/jordhy97/final_project}{https://github.com/jordhy97/final\_project}} The dataset consists of thousands of hotel reviews, which each contain a span label for aspect and sentiment words representing the opinion of the reviewer on the corresponding aspect. The labels use Inside-Outside-Beginning (IOB) tagging representation with two kinds of tags, aspect and sentiment.

\begin{table*}[!t]
\centering
\resizebox{1.0\textwidth}{!}{
\begin{tabular}{lrrrcl}
\toprule
\textbf{Dataset} & \textbf{\# Words} & \textbf{\# Sentences} & \textbf{Size} & \textbf{Style}& \textbf{Source} \\
\midrule
OSCAR~\cite{suarez2019oscar} & 2,279,761,186 & 148,698,472 & 14.9 GB & mixed & OSCAR \\
CoNLLu Common Crawl~\cite{conll2017commoncrawl} & 905,920,488 & 77,715,412 & 6.1 GB & mixed & LINDAT/CLARIAH-CZ \\
OpenSubtitles~\cite{Lison2016OpenSubtitles2016EL} & 105,061,204 & 25,255,662 & 664.8 MB & mixed & OPUS OpenSubtitles \\
Twitter Crawl$^2$ & 115,205,737 & 11,605,310 & 597.5 MB & colloquial & Twitter \\
Wikipedia Dump$^1$ & 76,263,857 & 4,768,444 & 528.1 MB & formal & Wikipedia \\
Wikipedia CoNLLu~\cite{conll2017commoncrawl} & 62,373,352 & 4,461,162 & 423.2 MB & formal & LINDAT/CLARIAH-CZ \\
Twitter UI$^2$~\cite{saputri2018emotion} & 16,637,641 & 1,423,212 & 88 MB & colloquial & Twitter \\
OPUS JW300~\cite{agic2019jw300} & 8,002,490 & 586,911 & 52 MB & formal & OPUS \\
Tempo$^3$ & 5,899,252 & 391,591 & 40.8 MB & formal & ILSP \\
Kompas$^3$ & 3,671,715 & 220,555 & 25.5 MB & formal & ILSP \\
TED & 1,483,786 & 111,759 & 9.9 MB & mixed & TED \\
BPPT & 500,032 & 25,943 & 3.5 MB & formal & BPPT \\
Parallel Corpus & 510,396 & 35,174 & 3.4 MB & formal & PAN Localization \\
TALPCo~\cite{nomoto2018tufs} & 8,795 & 1,392 & 56.1 KB & formal & Tokyo University \\
Frog Storytelling~\cite{moeljadi2014usage} & 1,545 & 177 & 10.1 KB & mixed & Tokyo University \\
\midrule
\textbf{TOTAL} & 3,581,301,476 & 275,301,176 & 23.43 GB & & \\
\bottomrule
\end{tabular}
}
\caption{\texttt{Indo4B} dataset statistics. $^1$ \hyperlink{https://dumps.wikimedia.org/backup-index.html}{https://dumps.wikimedia.org/backup-index.html}.  $^2$ We crawl tweets from Twitter. The Twitter data will not be shared publicly due to restrictions of the Twitter Developer Policy and Agreement. $^3$ \hyperlink{https://ilps.science.uva.nl/}{https://ilps.science.uva.nl/}. }
\label{tab:ID4B_corpus_stats}
\end{table*}

\paragraph{KEPS}
This keyphrase extraction dataset~\cite{mahfuzh2019improving} consists of text from Twitter discussing banking products and services and is written in the Indonesian language. A phrase containing important information is considered a keyphrase. Text may contain one or more keyphrases since important phrases can be located at different positions. The dataset follows the IOB chunking format, which represents the position of the keyphrase.  

\paragraph{NERGrit} This NER dataset is taken from the Grit-ID repository,\footnote{\hyperlink{https://github.com/grit-id/nergrit-corpus}{https://github.com/grit-id/nergrit-corpus}} and the labels are spans in IOB chunking representation. The dataset consists of three kinds of named entity tags, PERSON (name of person), PLACE (name of location), and ORGANIZATION (name of organization).

\paragraph{NERP}
This NER dataset ~\cite{hoesen2018investigating} contains texts collected from several Indonesian news websites. There are five labels available in this dataset, PER (name of person), LOC (name of location), IND (name of product or brand), EVT (name of the event), and FNB (name of food and beverage). Similar to the TermA dataset, the NERP dataset uses the IOB chunking format.

\subsubsection{Sentence-Pair Sequence Labeling Task}
\paragraph{FacQA} 
The goal of the FacQA dataset is to find the answer to a question from a provided short passage from a news article \cite{purwarianti2007qa}. Each row in the FacQA dataset consists of a question, a short passage, and a label phrase, which can be found inside the corresponding short passage. There are six categories of questions: date, location, name, organization, person, and quantitative. 


\subsection{\texttt{Indo4B} Dataset}
\label{sec:indo4b}

Indonesian NLP development has struggled with the availability of data. To cope with this issue, we provide a large-scale dataset called \texttt{Indo4B} for building a self-supervised pre-trained model. Our self-supervised dataset consists of around 4B words, with around 250M sentences. The \texttt{Indo4B} dataset covers both formal and colloquial Indonesian sentences compiled from 12 datasets, of which two cover Indonesian colloquial language, eight cover formal Indonesian language, and the rest have a mixed style of both colloquial and formal. The statistics of our large-scale dataset can be found in Table \ref{tab:ID4B_corpus_stats}. We share the datasets that are listed in the table, except for those from Twitter due to restrictions of the Twitter Developer Policy and Agreement. The details of \texttt{Indo4B} dataset sources are shown in Appendix B.

\begin{table*}[!t]
\centering
\resizebox{1.0\textwidth}{!}{
\begin{tabular}{lrrrcrrrc}
\toprule
\multirow{2}{*}{\textbf{Model}} & \multicolumn{4}{c}{\textbf{Maximum Sequence Length = 128}} & \multicolumn{4}{c}{\textbf{Maximum Sequence Length = 512}} \\ \cmidrule{2-9}
& \textbf{Batch Size} & \textbf{Learning Rate} & \textbf{Steps} & \textbf{Duration (Hr.)} & \textbf{Batch Size} & \textbf{Learning Rate} & \textbf{Steps} & \textbf{Duration (Hr.)} \\
\midrule
$\text{IndoBERT-lite}_{\text{BASE}}$ & 4096 & 0.00176 & 112.5 K & 38 & 1024 & 0.00088 & 50 K & 23 \\
$\text{IndoBERT}_{\text{BASE}}$ & 256 & 0.00002 & 1 M & 35 & 256 & 0.00002 & 68 K & 9 \\
$\text{IndoBERT-lite}_{\text{LARGE}}$ & 1024 & 0.00044 & 500 K & 134 & 256 & 0.00044 & 129 K & 45 \\
$\text{IndoBERT}_{\text{LARGE}}$ & 256 & 0.0001 & 1 M & 89 & 128 & 0.00008 & 120 K & 32 \\
\bottomrule
\end{tabular}
}
\caption{Hyperparameters and training duration for IndoBERT model pre-training.}
\label{}
\end{table*}

\subsection{Baselines}
\label{sec:baseline}
In this section, we explain the baseline models and the fine-tuning settings that we use in the \texttt{IndoNLU} benchmark.

\subsubsection{Models}
We provide a diverse set of baseline models, from a non-pre-trained model (scratch), to a word-embedding-based model, to contextualized language models. For the word-embeddings-based model, we use an existing fastText model trained on the Indonesian Common Crawl (CC-ID) dataset \cite{joulin2016fastText, grave2018learning}. 

\paragraph{fastText} We build a fastText model with our large-scale self-supervised dataset, \texttt{Indo4B}, for comparison with the CC-ID fastText model and contextualized language model. For the models above and the fastText model, we use the transformer architecture \cite{vaswani2017attention}. We experiment with different numbers of layers, 2, 4, and 6, for the transformer encoder. For the fastText model, we first pre-train the fastText embeddings with skipgram word representation and produce a 300-dimensional embedding vector. We then generate all required embeddings for each downstream task from the pre-trained fastText embeddings and cover all words in the vocabulary.

\paragraph{Contextualized Language Models} We build our own Indonesian BERT and ALBERT models, named IndoBERT and IndoBERT-lite, respectively, in both base and large sizes. The details of our IndoBERT and IndoBERT-lite models are explained in Section \ref{sec:IndoBERT}. Aside from a monolingual model, we also provide multilingual model baselines such as Multilingual BERT \cite{devlin2019bert}, XLM \cite{lample2019xlm}, and XLM-R \cite{conneau2019xlmr}. The details of each model are shown in Table ~\ref{tab:baseline-models}.

\subsubsection{Fine-tuning Settings}

We fine-tune a pre-trained model for each task with initial learning with a range of learning rates [1e-5, 4e-5]. We apply a decay rate of [0.8, 0.9] for every epoch, and sample each batch with a size of 16 for all datasets except FacQA and POSP, for which we use a batch size of 8. To establish a benchmark, we keep a fixed setting, and we use an early stop on the validation score to choose the best model. The details of the fine-tuning hyperparameter settings used are shown in Appendix D.

\subsection{Evaluation Metrics}
\label{sec:evaluation-metric}

We use the F1 score to measure the evaluation performance of all tasks. For the binary and multi-label classification tasks, we measure the macro-averaged F1 score by taking the top-1 prediction from the model. For the sequence labeling task, we calculate word-level sequence labeling macro-averaged F1-score for all models by following the sequence labeling evaluation method described in the CoNLL evaluation script. We calculate two mean F1-scores separately for classification and sequence labeling tasks to evaluate models on our \texttt{IndoNLU} benchmark.


\section{IndoBERT}
\label{sec:IndoBERT}

In this section, we describe the details of our Indonesian contextualized models, IndoBERT and IndoBERT-lite, which are trained using our \texttt{Indo4B} dataset. We elucidate the extensive details of the models' development, first the dataset preprocessing, followed by the pre-training setup.

\subsection{Preprocessing}

\paragraph{Dataset Preparation} To get the most beneficial next sentence prediction task training from the \texttt{Indo4B} dataset, we do either a paragraph separation or line separation if we notice document separator absence in the dataset. This document separation is crucial as it is used in the BERT architecture to extract long contiguous sequences ~\cite{devlin2019bert}. A separation between sentences with a new line is also required to differentiate each sentence. These are used by BERT to create input embeddings out of sentence pairs that are compacted into a single sequence. We specify the number of duplication factors for each of the datasets differently due to the various formats of the datasets that we collected.  We create duplicates on datasets with the end of document separators with a higher duplication factor. The preprocessing method is applied in both the IndoBERT and IndoBERT-lite models.

We keep the original form of a word to hold its contextual information since Indonesian words are built with rich morphological operations, such as compounding, affixation,  and reduplication~\cite{pisceldo2008two}. In addition, this setting is also suitable for contextual pre-training models that leverage inflections to improve the sentence-level representations.\cite{kutuzov-kuzmenko-2019-lemmatize}

Twitter data contains specific details, such as usernames, hashtags, emails, and URL hyperlinks. To preserve privacy and also to reduce noise, this private information in the Twitter UI dataset ~\cite{saputri2018emotion} is masked into generics tokens such as \texttt{<username>}, \texttt{<hashtag>}, \texttt{<email>} and \texttt{<links>}. On the other hand, this information is discarded in the larger Twitter Crawl dataset.

\paragraph{Vocabulary}
For both the IndoBERT and the IndoBERT-lite models, we utilize SentencePiece~\cite{kudo-richardson-2018-sentencepiece} with a byte pair encoding (BPE) tokenizer as the vocabulary generation method. We use a vocab size of 30.522 for the IndoBERT models and vocab size of 30.000 for the IndoBERT-lite models.

\begin{table*}[!t]
\centering
\resizebox{1.0\textwidth}{!}{
\begin{tabular}{lrrrrrr|rrrrrrrr}
\toprule 
\multirow{2}{*}{\textbf{Model}} & \multicolumn{6}{c}{\textbf{Classification}} & \multicolumn{8}{c}{\textbf{Sequence Labeling}} \\ \cmidrule{2-7} \cmidrule{8-15}
&  \textbf{EmoT} &  \textbf{SmSA} &  \textbf{CASA} &  \textbf{HoASA}    &  \textbf{WReTE} &  \textbf{AVG} & \textbf{POSP} & \textbf{BaPOS} &  \textbf{TermA}& \textbf{KEPS} &  \textbf{NERGrit} &  \textbf{NERP} &    \textbf{FacQA}   &  \textbf{AVG} \\ 
\midrule
Scratch & 57.31 & 67.35 & 67.15 & 76.28 & 64.35 & 66.49 & 86.78 & 70.24 & 70.36 & 39.40 & 5.80 & 30.66 & 5.00 & 44.03 \\
\midrule
fastText-cc-id & 65.36 & 76.92 & 79.02 & 85.32 & \underline{67.36} & 74.79 & 94.35 & 79.85 & \underline{76.12} & 56.39 & 37.32 & 46.46 & 15.29 & 57.97   \\
fastText-indo4b & \underline{69.23} & \underline{82.13} & \underline{82.20} & \underline{85.88} & 60.42 & \underline{75.97} & \underline{94.94} & \underline{81.77} & 74.43 & \underline{56.70} & \underline{38.69} & \underline{46.79} & 14.65 & \underline{58.28} \\
\midrule
mBERT & 67.30 & 84.14 & 72.23 & 84.63 & 84.40 & 78.54 & 91.85 & 83.25 & 89.51 & 64.31 & 75.02 & 69.27 & 61.29 & 76.36 \\
$\text{XLM-MLM}$ & 65.75 & 86.33 & 82.17 & 88.89 & 64.35 & 77.50 & \underline{\textbf{95.87}} & \underline{88.40} & 90.55 & 65.35 & 74.75 & 75.06 & 62.15 & 78.88 \\
$\text{XLM-R}_{\text{BASE}}$ & 71.15 & 91.39 & 91.71 & 91.57 & 79.95 & 85.15 & 95.16 & 84.64 & 90.99 & 68.82 & \underline{\textbf{79.09}} & 75.03 & 64.58 & 79.76 \\
$\text{XLM-R}_{\text{LARGE}}$ & \underline{78.51} & \underline{92.35} & \underline{92.40} & \underline{\textbf{94.27}} & \underline{83.82} & \underline{88.27} & 92.73 & 87.03 & \underline{91.45} & \underline{70.88} & 78.26 & \underline{78.52} & \underline{\textbf{74.61}} & \underline{\textbf{81.92}} \\
\midrule
${\text{IndoBERT-lite}_{\text{BASE}}}^{\dagger}$ & 73.88 & 90.85 & 89.68 & 88.07 & 82.17 & 84.93 & 91.40 & 75.10 & 89.29 & 69.02 & 66.62 & 46.58 & 54.99 & 70.43 \\
\hspace{3mm} + phase two & 72.27 & 90.29 & 87.63 & 87.62 & 83.62 & 84.29 & 90.05 & 77.59 & 89.19 & 69.13 & 66.71 & 50.52 & 49.18 & 70.34 \\
${\text{IndoBERT}_{\text{BASE}}}^{\dagger}$ & 75.48 & 87.73 & 93.23 & 92.07 & 78.55 & 85.41 & 95.26 & 87.09 & 90.73 & 70.36 & 69.87 & 75.52 & 53.45 & 77.47  \\
\hspace{3mm} + phase two & 76.28 & 87.66 & 93.24 & 92.70 & 78.68 & 85.71 & 95.23 & 85.72 & 91.13 & 69.17 & 67.42 & 75.68 & 57.06 & 77.34  \\
$\text{IndoBERT-lite}_{\text{LARGE}}$ & 75.19 & 88.66 & 90.99 & 89.53 & 78.98 & 84.67 & 91.56 & 83.74 & 90.23 & 67.89 & 71.19 & 74.37 & 65.50 & 77.78  \\
\hspace{3mm} + phase two & 70.80 & 88.61 & 88.13 & 91.05 & \underline{\textbf{85.41}} & 84.80 & 94.53 & 84.91 & 90.72 & 68.55 & 73.07 & 74.89 & 62.87 & 78.51 \\
$\text{IndoBERT}_{\text{LARGE}}$ & 77.08 & \underline{\textbf{92.72}} & \underline{\textbf{95.69}} & \underline{93.75} & 82.91 & \underline{\textbf{88.43}} & \underline{95.71} & \underline{\textbf{90.35}} & 91.87 & 71.18 & \underline{77.60} & \underline{\textbf{79.25}} & 62.48 & 81.21 \\
\hspace{3mm} + phase two & \underline{\textbf{79.47}} & 92.03 & 94.94 & 93.38 & 80.30 & 88.02 & 95.34 & 87.36 & \underline{\textbf{92.14}} & \underline{\textbf{71.27}} & 76.63 & 77.99 & \underline{68.09} & \underline{81.26} \\
\bottomrule
\end{tabular}
}
\caption{Results of baseline models with best performing configuration on the \texttt{IndoNLU} benchmark. Extensive experimental results are shown in Appendix E. Bold numbers are the best results among all. $^{\dagger}$The $\text{IndoBERT}$ models are trained using two training phases.}
\label{tab:benchmark-result}
\end{table*}

\subsection{Pre-training Setup}
All IndoBERT models are trained on TPUv3-8 in two phases. In the first phase, we train the models with a maximum sequence length of 128. The training takes around 35, 89, 38 and 134 hours on $\text{IndoBERT}_{\text{BASE}}$, $\text{IndoBERT}_{\text{LARGE}}$, $\text{IndoBERT-lite}_{\text{BASE}}$, and  $\text{IndoBERT-lite}_{\text{LARGE}}$, respectively. In the second phase, we continue the training of the IndoBERT models with a maximum sequence length of 512. It takes 9, 32, 23 and 45 hours on $\text{IndoBERT}_{\text{BASE}}$, $\text{IndoBERT}_{\text{LARGE}}$, $\text{IndoBERT-lite}_{\text{BASE}}$, and $\text{IndoBERT-lite}_{\text{LARGE}}$, respectively. The details of the pre-training hyperparameter settings are shown in Appendix D.

\paragraph{IndoBERT} We use a batch size of 256 and a learning rate of 2e-5 in both training phases for IndoBERT$_{\text{BASE}}$, and we adjust the learning rate to 1e-4 for IndoBERT$_{\text{LARGE}}$ to stabilize the training. Due to memory limitation, we scale down the batch size to 128 and the learning rate to 8e-5 in the second phase of the training, with a number of training steps adapted accordingly.
The base and large models are trained using the masked language modeling loss. We limit the maximum prediction per sequence into 20 tokens.

\paragraph{IndoBERT-lite} We follow the ALBERT pre-training hyperparameters setup~\cite{Lan2020ALBERT} to pre-train the IndoBERT-lite models. We limit the maximum prediction per sequence into 20 tokens on the models, pre-training with whole word masked loss. We train the base model with a batch size of 4096 in the first phase, and 1024 in the second phase. Since we have a limitation in computation power, we use a smaller batch size of 1024 in the first phase and 256 in the second phase in training our large model.

\section{Results and Analysis}

In this section, we show the results of the \texttt{IndoNLU} benchmark and analyze the performance of our models in terms of downstream tasks score and performance-space trade-off. In addition, we show an analysis of the effectiveness of using our collected data compared to existing baselines.

\subsection{Benchmark Results}
\paragraph{Overall Performance}
As mentioned in Section \ref{sec:indonlu}, we fine-tune all baseline models mentioned in Section \ref{sec:baseline}, and evaluate the model performance over all tasks, grouped into two categories, classification and sequence labeling. We can see in Table \ref{tab:benchmark-result}, that IndoBERT$_{\text{LARGE}}$, XLM-R$_{\text{LARGE}}$, and IndoBERT$_{\text{BASE}}$ achieve the top-3 best performance results on the classification tasks, and XLM-R$_{\text{LARGE}}$, IndoBERT$_{\text{LARGE}}$, and XLM-R$_{\text{BASE}}$ achieve the top-3 best performance results on the sequence labeling tasks. The experimental results also suggest that larger models have a performance advantage over smaller models. 
It is also evident that all pre-trained models outperform the scratch model, which shows the effectiveness of model pre-training. Another interesting observation is that all contextualized pre-trained models outperform word embeddings-based models by significant margins. This shows the superiority of the contextualized embeddings approach over the word embeddings approach. 

\begin{figure*}[t!]
    \centering
    \resizebox{0.99\textwidth}{!}{  
        \includegraphics{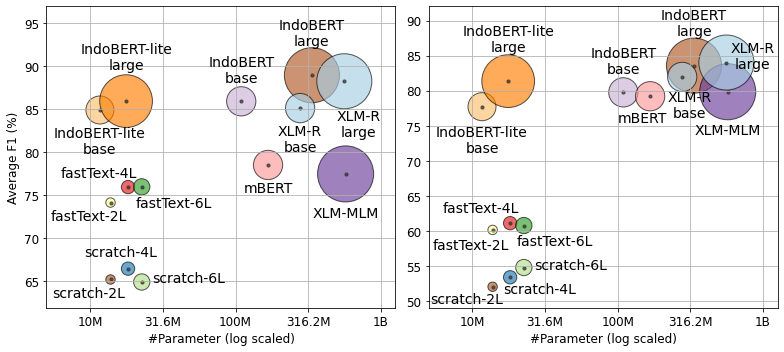}
    }
    \caption{Performance-space trade-off for all baseline models on classification tasks (left) and sequence labeling tasks (right). We take the best model for each model size. 2L, 4L, and 6L denote the number of layers used in the model. The size of the dots represents the number of FLOPs of the model.  We use python package \texttt{thop} taken from \hyperlink{https://pypi.org/project/thop/}{https://pypi.org/project/thop/} to calculate the number of FLOPs.}
    \label{fig:performance_speed}
\end{figure*}

\subsection{Performance-Space Trade-off}
Figure \ref{fig:performance_speed} shows the model performance with respect to the number of parameters. We can see two large clusters. On the bottom left, the scratch and fastText models appear, and they have the lowest F1 scores and the least floating points in the inference time. On the top right, we can see that the pre-trained models achieve decent performance, but in the inference time, they incur a high computation cost. Interestingly, in the top-left region, we can see the IndoBERT-lite models, which achieve similar performance to the IndoBERT models, but with many fewer parameters and a slightly lower computation cost.


\subsection{Multilingual vs. Monolingual Models}

Based on Table \ref{tab:benchmark-result}, we can conclude that contextualized monolingual models outperform contextualized multilingual models on the classification tasks by a large margin, but on the sequence labeling tasks, multilingual models tend to perform better compared to monolingual models and even perform much better on the NERGrit and FacQA tasks. As shown in Appendix A, both the NERGrit and FacQA tasks contain many entity names which come from other languages, especially English. These facts suggest that monolingual models capture the semantic meaning of a word better than multilingual models, but multilingual models identify foreign terms better than monolingual models.

\subsection{Effectiveness of \texttt{Indo4B} Dataset}

\begin{table}[h]
\centering
\resizebox{0.48\textwidth}{!}{
\begin{tabular}{lccc}
\toprule
\textbf{Tasks} & \textbf{\#Layer} & \textbf{fastText-cc-id} & \textbf{fastText-indo4b} \\
\midrule
\multirow{3}{*}{Classification}
& 2 & 72.00 &  \textbf{74.17} \\
& 4 & 74.79 &  \textbf{75.97} \\
& 6 & 74.80 &  \textbf{76.00} \\
\midrule
\multirow{3}{*}{\makecell{Sequence \\ Labeling}}
& 2 & \textbf{56.26} &  55.55 \\
& 4 & 57.97 &  \textbf{58.28} \\
& 6 & 56.82 &  \textbf{57.42} \\
\bottomrule
\end{tabular}
}
\caption{Experiment results on fastText embeddings on IndoNLU tasks with different number of transformer layers}

\label{tab:fasttext-result}
\end{table}

According to \citet{grave2018learning}, Common Crawl is a corpus containing over 24 TB.\footnote{\hyperlink{https://commoncrawl.github.io/cc-crawl-statistics/plots/languages}{https://commoncrawl.github.io/cc-crawl-statistics/plots/languages}} We estimate the size of the \texttt{CC-ID} dataset to be around $\approx$ 180 GB uncompressed. Although the \texttt{Indo4B} dataset size is much smaller ($\approx$ 23 GB), Table \ref{tab:fasttext-result} shows us that the fastText models trained on the \texttt{Indo4B} dataset (fastText-indo4b) consistently outperform fastText models trained on the \texttt{CC-ID} dataset (fastText-cc-id) in both classification and sequence labeling tasks in all model settings. Based on Table \ref{tab:benchmark-result}, the fact that fastText-indo4b outperforms fastText-cc-id with a higher score on 10 out of 12 tasks suggests that a relatively smaller dataset ($\approx$ 23 GB) can significantly outperform its larger counterpart ($\approx$ 180 GB). We conclude that even though our \texttt{Indo4B} dataset is smaller, it covers more variety of the Indonesian language and has better text quality compared to the \texttt{CC-ID} dataset.

\subsection{Effectiveness of IndoBERT and IndoBERT-lite}

Table \ref{tab:benchmark-result} shows that the IndoBERT models outperform the multilingual models on 8 out of 12 tasks. In general, the IndoBERT models achieve the highest average score on the classification task. We conjecture that monolingual models learn better sentiment-level semantics on both colloquial and formal language styles than multilingual models, even though the IndoBERT models' size is 40\%--60\% smaller. On sequence labeling tasks, the IndoBERT models cannot perform as well as the multilingual models (XLM-R) in three sequence labeling tasks: POSP, NERGrit, and FacQA. One of the possible explanations is that these datasets have many borrowed words from English, and multilingual models have the advantage in transferring learning from English. 

Meanwhile, the IndoBERT-lite models achieve a decent performance on both classification and sequence labeling tasks with the advantage of compact size. Interestingly, the $\text{IndoBERT-lite}_{\text{LARGE}}$ model performance is on par with that of $\text{XLM-R}_{\text{BASE}}$ while having 16x fewer parameters. We also observe that increasing the maximum sequence length to 512 in phase two improves the performance on the sequence labeling tasks. Moreover, training the model with longer input sequences enables it to learn temporal information from a given text input.

\section{Conclusion}
We introduce the first Indonesian benchmark for natural language understanding, \texttt{IndoNLU}, which consists of 12 tasks, with different levels of difficulty, domains, and styles. To establish a strong baseline, we collect large clean Indonesian datasets into a dataset called \texttt{Indo4B}, which we use for training monolingual contextual pre-trained language models, called IndoBERT and IndoBERT-lite. We demonstrate the effectiveness of our dataset and our pre-trained models in capturing sentence-level semantics, and apply them to the classification and sequence labeling tasks. To help with the reproducibility of the benchmark, we release the pre-trained models, including the collected data and code. In order to accelerate the community engagement and benchmark transparency, we have set up a leaderboard website for the NLP community. We publish our leaderboard website at \url{https://indobenchmark.com/}.

\section*{Acknowledgments}
We want to thank Cahya Wirawan, Pallavi Jain, Irene Gianni, Martijn Wieriks, Ade Romadhony, and Andrea Madotto for insightful discussions about this project. We sincerely thank the three anonymous reviewers for their insightful comments on our paper.



\bibliography{aacl-ijcnlp2020}

\begin{thebibliography}{36}
\expandafter\ifx\csname natexlab\endcsname\relax\def\natexlab#1{#1}\fi

\bibitem[{Agi{\'c} and Vuli{\'c}(2019)}]{agic2019jw300}
{\v{Z}}eljko Agi{\'c} and Ivan Vuli{\'c}. 2019.
\newblock Jw300: A wide-coverage parallel corpus for low-resource languages.
\newblock In \emph{Proceedings of the 57th Annual Meeting of the Association
  for Computational Linguistics}, pages 3204--3210.

\bibitem[{{Azhar} et~al.(2019){Azhar}, {Khodra}, and
  {Sutiono}}]{azhar2019aspectcategorization}
A.~N. {Azhar}, M.~L. {Khodra}, and A.~P. {Sutiono}. 2019.
\newblock Multi-label aspect categorization with convolutional neural networks
  and extreme gradient boosting.
\newblock In \emph{2019 International Conference on Electrical Engineering and
  Informatics (ICEEI)}, pages 35--40.

\bibitem[{Conneau et~al.(2019)Conneau, Khandelwal, Goyal, Chaudhary, Wenzek,
  Guzm{\'a}n, Grave, Ott, Zettlemoyer, and Stoyanov}]{conneau2019xlmr}
Alexis Conneau, Kartikay Khandelwal, Naman Goyal, Vishrav Chaudhary, Guillaume
  Wenzek, Francisco Guzm{\'a}n, Edouard Grave, Myle Ott, Luke Zettlemoyer, and
  Veselin Stoyanov. 2019.
\newblock Unsupervised cross-lingual representation learning at scale.
\newblock \emph{arXiv preprint arXiv:1911.02116}.

\bibitem[{Cui et~al.(2019)Cui, Che, Liu, Qin, Yang, Wang, and
  Hu}]{cui2019pretraining}
Yiming Cui, Wanxiang Che, Ting Liu, Bing Qin, Ziqing Yang, Shijin Wang, and
  Guoping Hu. 2019.
\newblock \href {http://arxiv.org/abs/1906.08101} {Pre-training with whole word
  masking for chinese bert}.

\bibitem[{Devlin et~al.(2019)Devlin, Chang, Lee, and
  Toutanova}]{devlin2019bert}
Jacob Devlin, Ming-Wei Chang, Kenton Lee, and Kristina Toutanova. 2019.
\newblock Bert: Pre-training of deep bidirectional transformers for language
  understanding.
\newblock In \emph{Proceedings of the 2019 Conference of the North American
  Chapter of the Association for Computational Linguistics: Human Language
  Technologies, Volume 1 (Long and Short Papers)}, pages 4171--4186.

\bibitem[{Dinakaramani et~al.(2014)Dinakaramani, Rashel, Luthfi, and
  Manurung}]{Dinakaramani2014}
Arawinda Dinakaramani, Fam Rashel, Andry Luthfi, and Ruli Manurung. 2014.
\newblock Designing an indonesian part of speech tagset and manually tagged
  indonesian corpus.
\newblock In \emph{2014 International Conference on Asian Language Processing,
  {IALP} 2014, Kuching, Malaysia, October 20-22, 2014}, pages 66--69. {IEEE}.

\bibitem[{Fernando et~al.(2019)Fernando, Khodra, and
  Septiandri}]{fern2019aspect}
Jordhy Fernando, Masayu~Leylia Khodra, and Ali~Akbar Septiandri. 2019.
\newblock \href {http://arxiv.org/abs/1908.04899} {Aspect and opinion terms
  extraction using double embeddings and attention mechanism for indonesian
  hotel reviews}.

\bibitem[{Ginter et~al.(2017)Ginter, Haji{\v c}, Luotolahti, Straka, and
  Zeman}]{conll2017commoncrawl}
Filip Ginter, Jan Haji{\v c}, Juhani Luotolahti, Milan Straka, and Daniel
  Zeman. 2017.
\newblock \href {http://hdl.handle.net/11234/1-1989} {{CoNLL} 2017 shared task
  - automatically annotated raw texts and word embeddings}.
\newblock {LINDAT}/{CLARIAH}-{CZ} digital library at the Institute of Formal
  and Applied Linguistics ({{\'U}FAL}), Faculty of Mathematics and Physics,
  Charles University.

\bibitem[{Grave et~al.(2018)Grave, Bojanowski, Gupta, Joulin, and
  Mikolov}]{grave2018learning}
Edouard Grave, Piotr Bojanowski, Prakhar Gupta, Armand Joulin, and Tomas
  Mikolov. 2018.
\newblock Learning word vectors for 157 languages.
\newblock In \emph{Proceedings of the International Conference on Language
  Resources and Evaluation (LREC 2018)}.

\bibitem[{Hoesen and Purwarianti(2018)}]{hoesen2018investigating}
Devin Hoesen and Ayu Purwarianti. 2018.
\newblock Investigating bi-lstm and crf with pos tag embedding for indonesian
  named entity tagger.
\newblock In \emph{2018 International Conference on Asian Language Processing
  (IALP)}, pages 35--38. IEEE.

\bibitem[{Ilmania et~al.(2018)Ilmania, Cahyawijaya, Purwarianti
  et~al.}]{ilmania2018aspect}
Arfinda Ilmania, Samuel Cahyawijaya, Ayu Purwarianti, et~al. 2018.
\newblock Aspect detection and sentiment classification using deep neural
  network for indonesian aspect-based sentiment analysis.
\newblock In \emph{2018 International Conference on Asian Language Processing
  (IALP)}, pages 62--67. IEEE.

\bibitem[{Joulin et~al.(2016)Joulin, Grave, Bojanowski, Douze, J{\'e}gou, and
  Mikolov}]{joulin2016fastText}
Armand Joulin, Edouard Grave, Piotr Bojanowski, Matthijs Douze, H{\'e}rve
  J{\'e}gou, and Tomas Mikolov. 2016.
\newblock Fasttext.zip: Compressing text classification models.
\newblock \emph{arXiv preprint arXiv:1612.03651}.

\bibitem[{Kudo and Richardson(2018)}]{kudo-richardson-2018-sentencepiece}
Taku Kudo and John Richardson. 2018.
\newblock \href {https://doi.org/10.18653/v1/D18-2012} {{S}entence{P}iece: A
  simple and language independent subword tokenizer and detokenizer for neural
  text processing}.
\newblock In \emph{Proceedings of the 2018 Conference on Empirical Methods in
  Natural Language Processing: System Demonstrations}, pages 66--71, Brussels,
  Belgium. Association for Computational Linguistics.

\bibitem[{Kurniawan and Aji(2018)}]{kurniawan2018toward}
Kemal Kurniawan and Alham~Fikri Aji. 2018.
\newblock Toward a standardized and more accurate indonesian part-of-speech
  tagging.
\newblock In \emph{2018 International Conference on Asian Language Processing
  (IALP)}, pages 303--307. IEEE.

\bibitem[{Kutuzov and Kuzmenko(2019)}]{kutuzov-kuzmenko-2019-lemmatize}
Andrey Kutuzov and Elizaveta Kuzmenko. 2019.
\newblock \href {https://www.aclweb.org/anthology/W19-6203} {To lemmatize or
  not to lemmatize: How word normalisation affects {ELM}o performance in word
  sense disambiguation}.
\newblock In \emph{Proceedings of the First NLPL Workshop on Deep Learning for
  Natural Language Processing}, pages 22--28, Turku, Finland. Link{\"o}ping
  University Electronic Press.

\bibitem[{Lample and Conneau(2019)}]{lample2019xlm}
Guillaume Lample and Alexis Conneau. 2019.
\newblock \href {http://arxiv.org/abs/1901.07291} {Cross-lingual language model
  pretraining}.
\newblock \emph{CoRR}, abs/1901.07291.

\bibitem[{Lan et~al.(2020)Lan, Chen, Goodman, Gimpel, Sharma, and
  Soricut}]{Lan2020ALBERT}
Zhenzhong Lan, Mingda Chen, Sebastian Goodman, Kevin Gimpel, Piyush Sharma, and
  Radu Soricut. 2020.
\newblock \href {https://openreview.net/forum?id=H1eA7AEtvS} {Albert: A lite
  bert for self-supervised learning of language representations}.
\newblock In \emph{International Conference on Learning Representations}.

\bibitem[{Le et~al.(2019)Le, Vial, Frej, Segonne, Coavoux, Lecouteux, Allauzen,
  Crabbé, Besacier, and Schwab}]{le2019flaubert}
Hang Le, Loïc Vial, Jibril Frej, Vincent Segonne, Maximin Coavoux, Benjamin
  Lecouteux, Alexandre Allauzen, Benoît Crabbé, Laurent Besacier, and Didier
  Schwab. 2019.
\newblock \href {http://arxiv.org/abs/1912.05372} {Flaubert: Unsupervised
  language model pre-training for french}.

\bibitem[{Lison and Tiedemann(2016)}]{Lison2016OpenSubtitles2016EL}
Pierre Lison and J{\"o}rg Tiedemann. 2016.
\newblock Opensubtitles2016: Extracting large parallel corpora from movie and
  tv subtitles.
\newblock In \emph{Proceedings of the 10th International Conference on Language
  Resources and Evaluation}.

\bibitem[{Liu et~al.(2019)Liu, Ott, Goyal, Du, Joshi, Chen, Levy, Lewis,
  Zettlemoyer, and Stoyanov}]{liu2019roberta}
Yinhan Liu, Myle Ott, Naman Goyal, Jingfei Du, Mandar Joshi, Danqi Chen, Omer
  Levy, Mike Lewis, Luke Zettlemoyer, and Veselin Stoyanov. 2019.
\newblock Roberta: A robustly optimized bert pretraining approach.
\newblock \emph{arXiv preprint arXiv:1907.11692}.

\bibitem[{Mahfuzh et~al.(2019)Mahfuzh, Soleman, and
  Purwarianti}]{mahfuzh2019improving}
Miftahul Mahfuzh, Sidik Soleman, and Ayu Purwarianti. 2019.
\newblock Improving joint layer rnn based keyphrase extraction by using
  syntactical features.
\newblock In \emph{2019 International Conference of Advanced Informatics:
  Concepts, Theory and Applications (ICAICTA)}, pages 1--6. IEEE.

\bibitem[{Martin et~al.(2019)Martin, Muller, Suárez, Dupont, Romary, Éric
  Villemonte de~la Clergerie, Seddah, and Sagot}]{martin2019camembert}
Louis Martin, Benjamin Muller, Pedro Javier~Ortiz Suárez, Yoann Dupont,
  Laurent Romary, Éric Villemonte de~la Clergerie, Djamé Seddah, and Benoît
  Sagot. 2019.
\newblock \href {http://arxiv.org/abs/1911.03894} {Camembert: a tasty french
  language model}.

\bibitem[{Moeljadi(2012)}]{moeljadi2014usage}
David Moeljadi. 2012.
\newblock Usage of indonesian possessive verbal predicates: a statistical
  analysis based on questionnaire and storytelling surveys.
\newblock In \emph{APLL-5 conference. SOAS, University of London}.

\bibitem[{Nomoto et~al.(2018)Nomoto, Okano, Moeljadi, and
  Sawada}]{nomoto2018tufs}
Hiroki Nomoto, Kenji Okano, David Moeljadi, and Hideo Sawada. 2018.
\newblock Tufs asian language parallel corpus (talpco).
\newblock In \emph{Proceedings of the Twenty-fourth Annual Meeting of the
  Association for Natural Language Processing}, pages 436--439.

\bibitem[{Ortiz~Su{\'a}rez et~al.(2019)Ortiz~Su{\'a}rez, Sagot, and
  Romary}]{suarez2019oscar}
Pedro~Javier Ortiz~Su{\'a}rez, Beno{\^i}t Sagot, and Laurent Romary. 2019.
\newblock \href {https://doi.org/10.14618/IDS-PUB-9021} {{Asynchronous Pipeline
  for Processing Huge Corpora on Medium to Low Resource Infrastructures}}.
\newblock In \emph{{7th Workshop on the Challenges in the Management of Large
  Corpora (CMLC-7)}}, Cardiff, United Kingdom. {Leibniz-Institut f{\"u}r
  Deutsche Sprache}.

\bibitem[{Peters et~al.(2018)Peters, Neumann, Iyyer, Gardner, Clark, Lee, and
  Zettlemoyer}]{peters2018deep}
Matthew Peters, Mark Neumann, Mohit Iyyer, Matt Gardner, Christopher Clark,
  Kenton Lee, and Luke Zettlemoyer. 2018.
\newblock Deep contextualized word representations.
\newblock In \emph{Proceedings of the 2018 Conference of the North American
  Chapter of the Association for Computational Linguistics: Human Language
  Technologies, Volume 1 (Long Papers)}, pages 2227--2237.

\bibitem[{Pisceldo et~al.(2008)Pisceldo, Mahendra, Manurung, and
  Arka}]{pisceldo2008two}
Femphy Pisceldo, Rahmad Mahendra, Ruli Manurung, and I~Wayan Arka. 2008.
\newblock A two-level morphological analyser for the indonesian language.
\newblock In \emph{Proceedings of the Australasian Language Technology
  Association Workshop 2008}, pages 142--150.

\bibitem[{Purwarianti and Crisdayanti(2019)}]{purwarianti2019improving}
Ayu Purwarianti and Ida Ayu Putu~Ari Crisdayanti. 2019.
\newblock Improving bi-lstm performance for indonesian sentiment analysis using
  paragraph vector.
\newblock In \emph{2019 International Conference of Advanced Informatics:
  Concepts, Theory and Applications (ICAICTA)}, pages 1--5. IEEE.

\bibitem[{Purwarianti et~al.(2007)Purwarianti, Tsuchiya, and
  Nakagawa}]{purwarianti2007qa}
Ayu Purwarianti, Masatoshi Tsuchiya, and Seiichi Nakagawa. 2007.
\newblock A machine learning approach for indonesian question answering system.
\newblock In \emph{Artificial Intelligence and Applications}, pages 573--578.

\bibitem[{Saputri et~al.(2018)Saputri, Mahendra, and
  Adriani}]{saputri2018emotion}
Mei~Silviana Saputri, Rahmad Mahendra, and Mirna Adriani. 2018.
\newblock Emotion classification on indonesian twitter dataset.
\newblock In \emph{2018 International Conference on Asian Language Processing
  (IALP)}, pages 90--95. IEEE.

\bibitem[{Septiandri and Sutiono(2019)}]{septi2019aspect}
Ali~Akbar Septiandri and Arie~Pratama Sutiono. 2019.
\newblock \href {http://arxiv.org/abs/1909.11879} {Aspect and opinion term
  extraction for aspect based sentiment analysis of hotel reviews using
  transfer learning}.

\bibitem[{Setya and Mahendra(2018)}]{ken2018entailment}
Ken~Nabila Setya and Rahmad Mahendra. 2018.
\newblock \href {https://doi.org/10.13140/RG.2.2.18820.27521} {Semi-supervised
  textual entailment on indonesian wikipedia data}.
\newblock In \emph{2018 International Conference on Computational Linguistics
  and Intelligent Text Processing (CICLing)}.

\bibitem[{Vaswani et~al.(2017)Vaswani, Shazeer, Parmar, Uszkoreit, Jones,
  Gomez, Kaiser, and Polosukhin}]{vaswani2017attention}
Ashish Vaswani, Noam Shazeer, Niki Parmar, Jakob Uszkoreit, Llion Jones,
  Aidan~N Gomez, {\L}ukasz Kaiser, and Illia Polosukhin. 2017.
\newblock Attention is all you need.
\newblock In \emph{Advances in neural information processing systems}, pages
  5998--6008.

\bibitem[{Wang et~al.(2019)Wang, Pruksachatkun, Nangia, Singh, Michael, Hill,
  Levy, and Bowman}]{wang2019superglue}
Alex Wang, Yada Pruksachatkun, Nikita Nangia, Amanpreet Singh, Julian Michael,
  Felix Hill, Omer Levy, and Samuel Bowman. 2019.
\newblock Superglue: A stickier benchmark for general-purpose language
  understanding systems.
\newblock In \emph{Advances in Neural Information Processing Systems}, pages
  3261--3275.

\bibitem[{Wang et~al.(2018)Wang, Singh, Michael, Hill, Levy, and
  Bowman}]{wang2018glue}
Alex Wang, Amanpreet Singh, Julian Michael, Felix Hill, Omer Levy, and Samuel
  Bowman. 2018.
\newblock Glue: A multi-task benchmark and analysis platform for natural
  language understanding.
\newblock In \emph{Proceedings of the 2018 EMNLP Workshop BlackboxNLP:
  Analyzing and Interpreting Neural Networks for NLP}, pages 353--355.

\bibitem[{Xu et~al.(2020)Xu, Zhang, Li, Hu, Cao, Liu, Li, Li, Sun, Xu
  et~al.}]{xu2020clue}
Liang Xu, Xuanwei Zhang, Lu~Li, Hai Hu, Chenjie Cao, Weitang Liu, Junyi Li,
  Yudong Li, Kai Sun, Yechen Xu, et~al. 2020.
\newblock Clue: A chinese language understanding evaluation benchmark.
\newblock \emph{arXiv preprint arXiv:2004.05986}.

\end{thebibliography}
\bibliographystyle{acl_natbib}

\appendix
\clearpage

\section{Data Samples}

In this section, we show examples for downstream tasks in the \texttt{IndoNLU} benchmark.

\begin{itemize}
    \item The examples of SmSA task are shown in Table \ref{task-SmSA}.
    \item The examples of EmoT task are shown in Table \ref{task-EmoT}.
    \item The examples of KEPS task are shown in Table \ref{task-KEPS}.
    \item The examples of HoASA task are shown in Table \ref{task-HoASA}. 
    \item The examples of CASA task are shown in Table \ref{task-CASA}.
    \item The examples of WReTE task are shown in Table \ref{task-WReTE}. 
    \item The examples of NERGrit task are shown in Table \ref{task-NERGrit}. 
    \item The examples of NERP task are shown in Table \ref{task-NERP}. 
    \item The examples of BaPOS task are shown in Table \ref{task-BaPOS}. 
    \item The examples of POSP task are shown in Table \ref{task-POSP}.
    \item The examples of FacQA task are shown in Table \ref{task-FacQA}. 
    \item The examples of TermA task are shown in Table \ref{task-TermA}.
\end{itemize}

\begin{table}[!b]
\centering
\resizebox{0.49\textwidth}{!}{
\begin{tabular}{ll}
\toprule
\textbf{Sentence} & \textbf{Sentiment} \\
\midrule
pengecut dia itu , cuma bisa nantangin dari belakang saja & neg \\
wortel mengandung vitamin a yang bisa jaga kesehatan mata & neut	\\
mocha float kfc itu minuman terenak yang pernah gue rasain & pos \\
\bottomrule
\end{tabular}
}
\caption{Sample data on task SmSA}
\label{task-SmSA}
\end{table}

\begin{table}[!b]
\centering
\resizebox{0.49\textwidth}{!}{
\begin{tabular}{ll}
\toprule
\textbf{Tweet} & \textbf{Emotion} \\
\midrule
Masalah ga akan pernah menjauh, hadapi Selasamu dengan penuh semangat! & happy \\
Sayang seribu sayang namun tak ada satupun yg nyangkut sampai sekarang & sadness \\
cewek suka bola itu dimata cowok cantiknya nambah, biarpun matanya panda & love \\
\bottomrule
\end{tabular}
}
\caption{Sample data on task EmoT}
\label{task-EmoT}
\end{table}

\begin{table}[!b]
\centering
\resizebox{0.49\textwidth}{!}{
\begin{tabular}{l|lllll}
\toprule
\textbf{Word} & Layanan & BCA & Mobile & Banking & Bermasalah \\
\textbf{Keyphrase} & O & B & I & I & B \\
\midrule
\textbf{Word} & Tidak & mengecewakan & pakai & BCA & Mobile	\\
\textbf{Keyphrase} & O & O & B & B & I \\
\midrule
\textbf{Word} & nggak & ada & tandingannya & e-channel & BCA\\
\textbf{Keyphrase} & B & I & I & B & I \\
\bottomrule
\end{tabular}
}
\caption{Sample data on task KEPS}
\label{task-KEPS}
\end{table}

\begin{table*}[!b]
\centering
\resizebox{0.95\textwidth}{!}{
\begin{tabular}{lllllllllll}
\toprule
\multirow{2}{*}{\textbf{Sentence}} & \multicolumn{10}{c}{\textbf{Aspect}} \\ 
\cmidrule{2-11}
& AC & Air Panas & Bau & General & Kebersihan & Linen & Service & Sunrise Meal & TV & WiFi \\
\midrule
air panas kurang berfungsi dan handuk lembab. & neut & neg & neut & neut & neut & neg & neut & neut & neut & neut \\
Shower zonk, resepsionis yang wanita judes & neut & neut & neut & neut & neut & neut & neg & neut & neut & neut \\
Kamar kurang bersih, terutama kamar mandi. & neut & neut & neut & neut & neg & neut & neut & neut & neut & neut \\
\bottomrule
\end{tabular}
}
\caption{Sample data on task HoASA}
\label{task-HoASA}
\end{table*}

\begin{table*}[!ht]
\centering
\resizebox{0.75\textwidth}{!}{
\begin{tabular}{lllllll}
\toprule
\multirow{2}{*}{\textbf{Sentence}} & \multicolumn{6}{c}{\textbf{Aspect}} \\ 
\cmidrule{2-7}
& Fuel & Machine & Others & Part & Price & Service \\
\midrule
bodi plus tampilan nya Avanza baru mantap juragan  & neut & neut & neut & pos & neut & neut \\
udah gaya nya stylish ekonomis pula, beli calya deh & neut & neut & neut & pos & pos & neut  \\
Mobil kualitas jelek kayak wuling saja masuk Indonesia &  neut &  neut &  neg & neut & neut  &  neut \\
\bottomrule
\end{tabular}
}
\caption{Sample data on task CASA}
\label{task-CASA}
\end{table*}

\begin{table*}[!t]
\centering
\resizebox{0.95\textwidth}{!}{
\begin{tabular}{lll}
\toprule
\textbf{Sentence A} & \textbf{Sentence B} & \textbf{Label} \\
\midrule
Anak sebaiknya menjalani tirah baring & Anak sebaiknya menjalani istirahat  & Entail or Paraphrase	\\
Kedua kata ini ditulis dengan huruf kanji yang sama  & Jepang disebut Nippon atau Nihon dalam bahasa Jepang   & Not Entail	\\
Elektron hanya menduduki 0,06\% massa total atom & Elektron hanya mengambil 0,06\% massa total atom & Entail or Paraphrase \\
\bottomrule
\end{tabular}
}
\caption{Sample data on task WReTE}
\label{task-WReTE}
\end{table*}

\begin{table*}[!ht]
\centering
\resizebox{0.95\textwidth}{!}{
\begin{tabular}{l|llllllllll}
\toprule
\textbf{Word} & Produser & David & Heyman & dan & sutradara & Mark & Herman &sedang & mencari & seseorang \\
\textbf{Entity} & O & B-PER & I-PER & O & O & B-PER & I-PERS & O & O & O \\
\midrule
\textbf{Word} & Pada & tahun & 1996 & Williams & pindah & ke & Sebastopol & , & California & di	\\
\textbf{Entity} & O & O & O & B-PER & O & O & B-PLA & O & B-PLA & O \\
\midrule
\textbf{Word} & bekerja & untuk & penerbitan & perusahaan & teknologi & O & , & Reilly & Media & .\\
\textbf{Entity} & O & O & O & O & O & B-ORG & I-ORG & I-ORG & I-ORG & O \\
\bottomrule
\end{tabular}
}
\caption{Sample data on task NERGrit. PER = PERSON, ORG = ORGANIZATION, PLA = PLACE}
\label{task-NERGrit}
\end{table*}

\begin{table*}[!ht]
\centering
\resizebox{0.95\textwidth}{!}{
\begin{tabular}{l|llllllllll}
\toprule
\textbf{Word} & kepala & dinas & tata & kota & manado & amos & kenda & menyatakan & tidak & tahu \\
\textbf{Entity} & O & O & O & O & B-PLC & B-PPL & I-PPL & O & O & O \\
\midrule
\textbf{Word} & telah & mendaftar & untuk & menjadi & official & merchant & bandung & great & sale & 2017	\\
\textbf{Entity} & O & O & O & O & O & O & B-EVT & I-EVT & I-EVT & I-EVT \\
\midrule
\textbf{Word} & sekitar & timur & dan & barat & arnhem & , & katherine & dan & daerah & sekitar \\
\textbf{Entity} & O & B-PLC & O & B-PLC & I-PLC & O & B-PLC & O & O & O  \\
\bottomrule
\end{tabular}
}
\caption{Sample data on task NERP. PLC = PLACE, PPL = PEOPLE, EVT = EVENT}
\label{task-NERP}
\end{table*}

\begin{table*}[!ht]
\centering
\resizebox{0.95\textwidth}{!}{
\begin{tabular}{l|llllllllll}
\toprule
\textbf{Word} & Pemerintah & kota & Delhi & mengerahkan & monyet & untuk & mengusir & monyet-monyet & lain & yang \\
\textbf{Tag} & B-NNP & B-NNP & B-NNP & B-VB & B-NN & B-SC & B-VB & B-NN & B-JJ & B-SC \\
\midrule
\textbf{Word} & Beberapa & laporan & menyebutkan & setidaknya & 10 & monyet & ditempatkan & di & luar & arena	\\
\textbf{Tag} & B-CD & B-NN & B-VB
 & B-RB & B-CD & B-NN & B-VB & B-IN & B-NN & B-NN \\
\midrule
\textbf{Word} & berencana & mendatangkan & 10 & monyet & sejenis & dari & negara & bagian & Rajasthan & . \\
\textbf{Tag} & B-VB & B-VB & B-CD & B-NN & B-NN & B-IN & B-NNP & I-NNP & B-NNP & B-Z \\
\bottomrule
\end{tabular}
}
\caption{Sample data on task BaPOS. POS tag labels follow Universitas Indonesia POS Tag Standard. \footnote{http://bahasa.cs.ui.ac.id/postag/downloads/Tagset.pdf}}
\label{task-BaPOS}
\end{table*}

\begin{table*}[!ht]
\centering
\resizebox{0.95\textwidth}{!}{
\begin{tabular}{l|llllllllll}
\toprule
\textbf{Word} & kepala & dinas & tata & kota & manado & amos & kenda & menyatakan & tidak & tahu \\
\textbf{Tag} & B-NNO & B-VBP & B-NNO & B-NNO & B-NNP & B-NNP & B-NNP & B-VBT & B-NEG & B-VBI \\
\midrule
\textbf{Word} & telah & mendaftar & untuk & menjadi & official & merchant & bandung & great & sale & 2017	\\
\textbf{Tag} & B-ADK & B-VBI & B-PPO & B-VBL & B-NNO & B-NNP & B-NNP & B-NNP & B-NNP & B-NUM \\
\midrule
\textbf{Word} & sekitar & timur & dan & barat & arnhem & , & katherine & dan & daerah & sekitar \\
\textbf{Tag} & B-PPO & B-NNP & B-CCN & B-NNP & B-NNP & B-SYM & B-NNP & B-CCN & B-NNO & B-ADV \\
\bottomrule
\end{tabular}
}
\caption{Sample data on task POSP POS tag labels follow INACL POS Tagging Convention. \footnote{http://inacl.id/inacl/wp-content/uploads/2017/06/INACL-POS-Tagging-Convention-26-Mei.pdf}}
\label{task-POSP}
\end{table*}

\begin{table*}[!ht]
\centering
\resizebox{0.95\textwidth}{!}{
\begin{tabular}{l|lllllll}
\toprule
\textbf{Question} & \multicolumn{7}{l}{"Siapakah penasihat utama Presiden AS George W Bush?"} \\
\textbf{Passage} & Nasib & Karl & Rove & Akan & Segera & Diputuskan & \\
\textbf{Label} & O & B & I & O & O & O & \\
\midrule
\textbf{Question} & \multicolumn{7}{l}{"Dimana terjadinya letusan gunung berapi dahsyat tahun 1883?"} \\
\textbf{Passage} & Di & Kepulauan & Krakatau & Terdapat & 400 & Tanaman & \\
\textbf{Label} & O & B & I & O & O & O & \\
\midrule
\textbf{Question} & \multicolumn{7}{l}{"Perusahaan apakah yang sejak 1 Januari 2006, menurunkan harga pertamax dan pertamax plus?"} \\
\textbf{Passage} & Pesaing & Semakin & Banyak & , &  Pertamina & Berusaha & Kompetitif \\
\textbf{Label} & O & O & O & O & B & O & O \\
\bottomrule
\end{tabular}
}
\caption{Sample data on task FacQA}
\label{task-FacQA}
\end{table*}

\begin{table*}[!ht]
\centering
\resizebox{0.95\textwidth}{!}{
\begin{tabular}{l|llllllllll}
\toprule
\textbf{Word} & sayang & wifi & tidak & bagus & harus & keluar & kamar & . & fasilitas & lengkap \\
\textbf{Entity} & O & B-ASP & B-SEN & I-SEN & O & O & O & O & B-ASP & B-SEN \\
\midrule
\textbf{Word} & pelayanan & nya & sangat & bagus & . & kamar & nya & juga & oke & .	\\
\textbf{Entity} & B-ASP & I-ASP & B-SEN & I-SEN & O & B-ASP & I-ASP & O & B-SEN & O \\
\midrule
\textbf{Word} & kamar & cukup & luas & , & interior & menarik & dan & unik & sekali & , \\
\textbf{Entity} & B-ASP & B-SEN & I-SEN & O & B-ASP & B-SEN & O & B-SEN & I-SEN & O  \\
\bottomrule
\end{tabular}
}
\caption{Sample data on task TermA. SEN = SENTIMENT, ASP = ASPECT}
\label{task-TermA}
\end{table*}

\section{\texttt{Indo4B} Data Sources}
In this section, we show the source of each dataset that we use to build our \texttt{Indo4B} dataset. The source of each corpus is shown in Table \ref{id4b-corpus}.

\begin{table*}[!ht]
\centering
\resizebox{0.95\textwidth}{!}{
\begin{tabular}{lll}
\toprule
\textbf{Corpus Name} & \textbf{Source} & \textbf{Public URL}\\
\midrule
OSCAR & OSCAR & https://oscar-public.huma-num.fr/compressed/id\_dedup.txt.gz \\
CoNLLu Common Crawl & LINDAT/CLARIAH-CZ & https://lindat.mff.cuni.cz/repository/xmlui/bitstream/handle/11234/1-1989/Indonesian-annotated-conll17.tar \\
OpenSubtitles & OPUS OpenSubtitles & http://opus.nlpl.eu/download.php?f=OpenSubtitles/v2016/mono/OpenSubtitles.raw.id.gz \\
Wikipedia Dump & Wikipedia & https://dumps.wikimedia.org/idwiki/20200401/idwiki-20200401-pages-articles-multistream.xml.bz2 \\
Wikipedia CoNLLu & LINDAT/CLARIAH-CZ & https://lindat.mff.cuni.cz/repository/xmlui/bitstream/handle/11234/1-1989/Indonesian-annotated-conll17.tar \\
Twitter Crawl & Twitter & Not publicly available \\
Twitter UI & Twitter & Not publicly available  \\
OPUS JW300 & OPUS & http://opus.nlpl.eu/JW300.php \\
Tempo & ILSP & http://ilps.science.uva.nl/ilps/wp-content/uploads/sites/6/files/bahasaindonesia/tempo.zip \\
Kompas & ILSP & http://ilps.science.uva.nl/ilps/wp-content/uploads/sites/6/files/bahasaindonesia/kompas.zip \\
TED & TED & https://github.com/ajinkyakulkarni14/TED-Multilingual-Parallel-Corpus/tree/master/Monolingual\_data \\
BPPT & BPPT & http://www.panl10n.net/english/outputs/Indonesia/BPPT/0902/BPPTIndToEngCorpusHalfM.zip \\
Parallel Corpus & PAN Localization & http://panl10n.net/english/outputs/Indonesia/UI/0802/Parallel/\%20Corpus.zip \\
TALPCo & Tokyo University & https://github.com/matbahasa/TALPCo \\
Frog Storytelling & Tokyo University & https://github.com/davidmoeljadi/corpus-frog-storytelling \\
\bottomrule
\end{tabular}
}
\caption{\texttt{Indo4B} Corpus}
\label{id4b-corpus}
\end{table*}

\section{Pre-Training Hyperparameters}
In this section, we show all hyperparameters used in our IndoBERT and IndoBERT-lite training process. The hyperparameters is shown in Table \ref{indobert-config}.

\begin{table*}[!ht]
\centering
\resizebox{0.85\textwidth}{!}{
\begin{tabular}{lrrrr}
\toprule
\textbf{Hyperparameter} & $\text{IndoBERT}_{\text{BASE}}$ & $\text{IndoBERT}_{\text{LARGE}}$ & $\text{IndoBERT-lite}_{\text{BASE}}$ & $\text{IndoBERT-lite}_{\text{LARGE}}$ \\
\midrule
attention\_probs\_dropout\_prob & 0.1 & 0.1 & 0 & 0 \\
hidden\_act & gelu & gelu & gelu & gelu  \\
hidden\_dropout\_prob & 0.1 & 0.1 & 0 & 0  \\
embedding\_size & 768 & 1024 & 128 & 128 \\
hidden\_size & 768 & 1024 & 768 & 1024 \\
initializer\_range & 0.02 & 0.02 & 0.02 & 0.02 \\
intermediate\_size & 3072 & 4096 & 3072 & 4096 \\
max\_position\_embeddings & 512 & 512 & 512 & 512 \\
num\_attention\_heads & 12 & 16 & 12 & 16 \\
num\_hidden\_layers & 12 & 24 & 12 & 24 \\
type\_vocab\_size & 2 & 2  & 2 & 2 \\
vocab\_size & 30522 & 30522 & 30000 & 30000 \\
num\_hidden\_groups & - & - & 1 & 1 \\
net\_structure\_type & - & -  & 0 & 0 \\
gap\_size & - & - & 0 & 0 \\
num\_memory\_blocks & - & - & 0 & 0 \\
inner\_group\_num & - & - & 1 & 1 \\
down\_scale\_factor & - & - & 1 & 1 \\
\bottomrule
\end{tabular}
}
\caption{Hyperparameter configurations for IndoBERT and IndoBERT-lite pre-trained models.}
\label{indobert-config}
\end{table*}

\section{Fine-Tuning Hyperparameters}
In this section, we show all hyperparameters used in the fine-tuning process of each baseline model. The hyperparameter configuration is shown in Table \ref{finetuning-config}.

\begin{table*}[!ht]
\centering
\resizebox{0.95\textwidth}{!}{
\begin{tabular}{lcccccccc} \toprule
 & \textbf{batch\_size} & \textbf{n\_layers} & \textbf{n\_epochs} & \textbf{lr} & \textbf{early\_stop} & \textbf{gamma} & \textbf{max\_norm} & \textbf{seed} \\ \midrule
Scratch & {[8,16]} & {[}2,4,6{]} & 25 & 1e-4 & 12 & 0.9 & 10 & 42 \\
fastText-cc-id & {[8,16]} & {[}2,4,6{]} & 25 & 1e-4 & 12 & 0.9 & 10 & 42 \\
fastText-indo4B & {[8,16]} & {[}2,4,6{]} & 25 & 1e-4 & 12 & 0.9 & 10 & 42 \\ \midrule
mBERT & {[8,16]} & 12 & 25 & 1e-5 & 12 & 0.9 & 10 & 42 \\
XLM-MLM & {[8,16]} & 16 & 25 & 1e-5 & 12 & 0.9 & 10 & 42 \\
$\text{XLM-R}_{\text{BASE}}$ & {[8,16]} & 12 & 25 & 2e-5 & 12 & 0.9 & 10 & 42 \\
$\text{XLM-R}_{\text{LARGE}}$ & {[8,16]} & 24 & 25 & 1e-5 & 12 & 0.9 & 10 & 42 \\ \midrule
$\text{IndoBERT-lite}_{\text{BASE}}$ & {[8,16]} & 12 & 25 & 1e-5 & 12 & 0.9 & 10 & 42 \\
\hspace{3mm}+ phase 2 & {[8,16]} & 12 & 25 & 1e-5 & 12 & 0.9 & 10 & 42 \\
$\text{IndoBERT-lite}_{\text{LARGE}}$ & {[8,16]} & 24 & 25 & [1e-5,2e-5] & 12 & 0.9 & 10 & 42 \\
\hspace{3mm}+ phase 2 & {[8,16]} & 24 & 25 & 2e-5 & 12 & 0.9 & 10 & 42 \\
$\text{IndoBERT}_{\text{BASE}}$ & {[8,16]} & 12 & 25 & [1e-5,4e-5] & 12 & 0.9 & 10 & 42 \\
\hspace{3mm}+ phase 2 & {[8,16]} & 12 & 25 & 4e-5 & 12 & 0.9 & 10 & 42 \\
$\text{IndoBERT}_{\text{LARGE}}$ & {[8,16]} & 24 & 25 & 4e-5 & 12 & 0.9 & 10 & 42 \\
\hspace{3mm}+ phase 2 & {[8,16]} & 24 & 25 & [3e-5,4e-5] & 12 & 0.9 & 10 & 42 \\ \bottomrule
\end{tabular}
}
\caption{Hyperparameter configurations for fine-tuning in \texttt{IndoNLU} benchmark.  We use a batch size of 8 for POSP and FacQA, and a batch size of 16 for EmoT, SmSA, CASA, HoASA, WReTE,  BaPOS, TermA, KEPS, NERGrit, and NERP.}
\label{finetuning-config}
\end{table*}

\section{Extensive Experiment Results on \texttt{IndoNLU} Benchmark}

In this section, we show all experiments conducted in the \texttt{IndoNLU} benchmark. We use a batch size of 16 for all datasets except FacQA and POSP, for which we use a batch size of 8. The results of the full experiments are shown in Table \ref{indonlu-results-all}.

\begin{table*}[!t]
\centering
\resizebox{0.99\textwidth}{!}{
\begin{tabular}{llrrrrrrrr|rrrrrrrr}
\toprule
\multirow{2}{*}{\textbf{Model}} & \multirow{2}{*}{\textbf{LR}} & \multirow{2}{*}{\textbf{\# Layer}} & \multirow{2}{*}{\textbf{Param}} & \multicolumn{6}{c|}{\textbf{Classification}} & \multicolumn{8}{c}{\textbf{Sequence Labeling}} \\
\cmidrule{5-18}
& & & & \textbf{EmoT} &  \textbf{SmSA} &  \textbf{CASA} &  \textbf{HoASA} &  \textbf{WReTE} &  \textbf{AVG} &  \textbf{POSP} &  \textbf{BaPOS} &  \textbf{TermA} &  \textbf{KEPS} &  \textbf{NERGrit} &  \textbf{NERP} &  \textbf{FacQA} &  \textbf{AVG} \\
\midrule
scratch & 1e-4 &  2 &   38.6M & 58.51 & 64.22 & 65.58 & 78.31 & 59.54 & 65.23 & 85.69 & 66.30 & 69.67 & 47.71 & 4.62 & 31.14 & 4.08 & 44.17 \\
scratch & 1e-4 &  4 &   52.8M &  57.31 & 67.35 & 67.15 & 76.28 & 64.35 & 66.49 & 86.78 & 70.24 & 70.36 & 39.40 & 5.80 & 30.66 & 5.00 & 44.03 \\
scratch & 1e-4 &  6 &   67.0M &  52.84 & 67.07 & 69.88 & 76.83 & 58.06 & 64.94 & 86.16 & 68.18 & 70.64 & 45.65 & 5.14 & 27.88 & 5.21 & 44.12 \\
\midrule
fasttext-cc-id-300-no-oov-uncased & 1e-4 &  6 &   15.1M &  67.43 & 78.84 & 81.61 & 85.01 & 61.13 & 74.80 & 94.36 & 78.45 & 77.26 & 57.28 & 26.70 & 46.36 & 17.3 & 56.82 \\
fasttext-cc-id-300-no-oov-uncased & 1e-4 &  4 &   10.7M &  65.36 & 76.92 & 79.02 & 85.32 & 67.36 & 74.79 & 94.35 & 79.85 & 76.12 & 56.39 & 37.32 & 46.46 & 15.29 & 57.97 \\
fasttext-cc-id-300-no-oov-uncased & 1e-4 &  2 &    6.3M &  64.74 & 76.71 & 75.39 & 78.05 & 65.11 & 72.00 & 94.42 & 78.12 & 73.45 & 55.22 & 33.27 & 45.44 & 13.89 & 56.26 \\
\midrule
fasttext-4B-id-300-no-oov-uncased & 1e-4 &  6 &   15.1M &  68.47 & 83.07 & 81.96 & 86.20 & 60.33 & 76.00 & 95.15 & 80.61 & 75.26 & 44.71 & 40.83 & 47.02 & 18.39 & 57.42 \\
fasttext-4B-id-300-no-oov-uncased & 1e-4 &  4 &   10.7M &  69.23 & 82.13 & 82.20 & 85.88 & 60.42 & 75.97 & 94.94 & 81.77 & 74.43 & 56.70 & 38.69 & 46.79 & 14.65 & 58.28 \\
fasttext-4B-id-300-no-oov-uncased & 1e-4 &  2 &    6.3M &  70.97 & 83.63 & 78.97 & 80.16 & 57.11 & 74.17 & 94.93 & 80.11 & 71.92 & 56.67 & 31.46 & 45.08 & 8.65 & 55.55 \\
\midrule
indobert-lite-base-128-112.5k & 1e-5 & 12 &   11.7M &  73.88 & 90.85 & 89.68 & 88.07 & 82.17 & 84.93 & 91.40 & 75.10 & 89.29 & 69.02 & 66.62 & 46.58 & 54.99 & 70.43 \\
indobert-lite-base-128-191.5k & 1e-5 & 12 &   11.7M &  71.95 & 89.87 & 84.71 & 87.57 & 80.30 & 82.88 & 87.27 & 67.33 & 89.15 & 65.84 & 67.67 & 49.32 & 51.76 & 68.33 \\
indobert-lite-base-512-162.5k & 1e-5 & 12 &   11.7M &  72.27 & 90.29 & 87.63 & 87.62 & 83.62 & 84.29 & 90.05 & 77.59 & 89.19 & 69.13 & 66.71 & 50.52 & 49.18 & 70.34 \\
\midrule
indobert-base-128 & 4e-5 & 12 &  124.5M &  75.48 & 87.73 & 93.23 & 92.07 & 78.55 & 85.41 & 95.26 & 87.09 & 90.73 & 70.36 & 69.87 & 75.52 & 53.45 & 77.47  \\
indobert-base-512 & 1e-5 & 12 &  124.5M &  76.61 & 90.90 & 91.77 & 90.70 & 79.73 & 85.94 & 95.10 & 86.25 & 90.58 & 69.39 & 63.67 & 75.36 & 53.14 & 76.21 \\
indobert-base-512 & 4e-5 & 12 &  124.5M &  76.28 & 87.66 & 93.24 & 92.70 & 78.68 & 85.71 & 95.23 & 85.72 & 91.13 & 69.17 & 67.42 & 75.68 & 57.06 & 77.34 \\
\midrule
indobert-lite-large-128 & 1e-5 & 24 &   17.7M &  75.19 & 88.66 & 90.99 & 89.53 & 78.98 & 84.67 & 91.56 & 83.74 & 90.23 & 67.89 & 71.19 & 74.37 & 65.50 & 77.78 \\
indobert-lite-large-512 & 1e-5 & 24 &   17.7M &  71.67 & 90.13 & 88.88 & 88.80 & 81.19 & 84.13 & 91.53 & 83.51 & 90.07 & 67.36 & 73.27 & 74.34 & 69.47 & 78.51 \\
indobert-lite-large-512 & 2e-5 & 24 &   17.7M &  70.80 & 88.61 & 88.13 & 91.05 & 85.41 & 84.80 & 94.53 & 84.91 & 90.72 & 68.55 & 73.07 & 74.89 & 62.87 & 78.51 \\
\midrule
indobert-large-128-1100k & 4e-5 & 24 &  335.2M &  77.04 & 93.71 & 96.64 & 93.27 & 84.17 & 88.97 & 95.71 & 89.74 & 91.97 & 70.82 & 70.76 & 77.54 & 67.27 & 80.55 \\
indobert-large-128-1000k & 4e-5 & 24 &  335.2M &  77.08 & 92.72 & 95.69 & 93.75 & 82.91 & 88.43 & 95.71 & 90.35 & 91.87 & 71.18 & 77.60 & 79.25 & 62.48 & 81.21 \\
indobert-large-512-1100k & 4e-5 & 24 &  335.2M &  77.39 & 92.90 & 95.90 & 93.77 & 81.62 & 88.32 & 95.25 & 86.05 & 91.92 & 69.71 & 75.20 & 77.53 & 69.86 & 80.79 \\
indobert-large-512-1100k & 3e-5 & 24 &  335.2M &  79.47 & 92.03 & 94.94 & 93.38 & 80.30 & 88.02 & 95.34 & 87.36 & 92.14 & 71.27 & 76.63 & 77.99 & 68.09 & 81.26 \\
\midrule
bert-base-multilingual-uncased & 1e-5 & 12 &  167.4M &  67.30 & 84.14 & 72.23 & 84.63 & 84.40 & 78.54 & 91.85 & 83.25 & 89.51 & 64.31 & 75.02 & 69.27 & 61.29 & 76.36 \\
xlm-mlm-100-1280 & 1e-5 & 16 &  573.2M &  65.75 & 86.33 & 82.17 & 88.89 & 64.35 & 77.50 & 95.87 & 88.40 & 90.55 & 65.35 & 74.75 & 75.06 & 62.15 & 78.88 \\
xlm-roberta-base & 2e-5 & 12 &  278.7M &  71.15 & 91.39 & 91.71 & 91.57 & 79.95 & 85.15 & 95.16 & 84.64 & 90.99 & 68.82 & 79.09 & 75.03 & 64.58 & 79.76 \\
xlm-roberta-large & 1e-5 & 24 &  561.0M &  78.51 & 92.35 & 92.40 & 94.27 & 83.82 & 88.27 & 92.73 & 87.03 & 91.45 & 70.88 & 78.26 & 78.52 & 74.61 & 81.92 \\
\bottomrule

\end{tabular}
}
\caption{Results of all experiments conducted in \texttt{IndoNLU} benchmark. We sample each batch with a size of 16 for all datasets except FacQA and POSP, for which we use a batch size of 8.}
\label{indonlu-results-all}
\end{table*}

\end{document}